\documentclass[11pt]{article}

\usepackage{acl}
\usepackage{times}
\usepackage{latexsym}

\usepackage[T1]{fontenc}

\usepackage[utf8]{inputenc}

\usepackage{microtype}

\usepackage{inconsolata}

\usepackage{graphicx}

\usepackage{booktabs}
\usepackage{amsmath}

\usepackage[most]{tcolorbox}
\usepackage{tabularx}
\usepackage{xcolor}

\usepackage{listings}
\usepackage{algorithm}
\usepackage{algpseudocode}
\usepackage{amsmath}

\usepackage{booktabs}
\usepackage{multirow}
\usepackage{xcolor}
\usepackage{adjustbox}

\usepackage{booktabs}
\usepackage{multirow}
\usepackage{colortbl}
\usepackage{xcolor}
\usepackage{adjustbox}

\usepackage{multirow}

\usepackage{booktabs}
\usepackage{multirow}
\usepackage{adjustbox}
\usepackage{xcolor}

\usepackage{placeins}

\title{\raisebox{-1.5ex}{\includegraphics[height=1.8em]{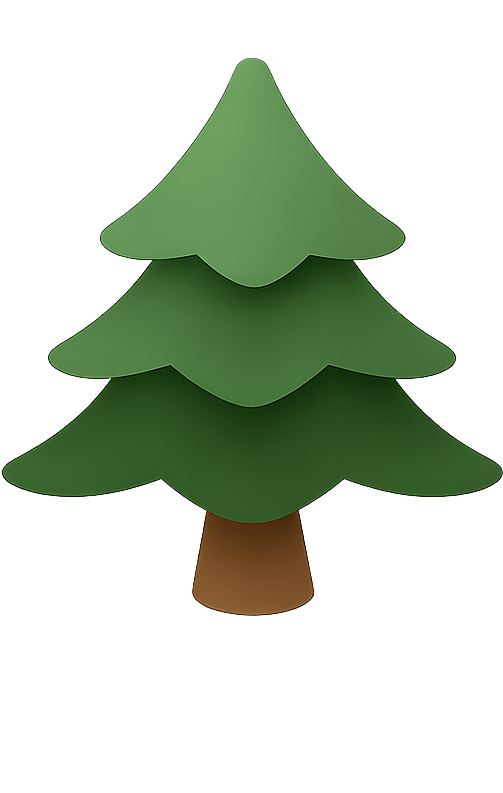}}\hspace{0.2em}\normalfont\bfseries AgriAgent: Contract-Driven Planning and Capability-Aware Tool Orchestration in Real-World Agriculture}

\author{
\textbf{Bo Yang$^{*}$, Yu Zhang$^{*}$, Yunkui Chen, Lanfei Feng} \\[3.5pt]
\textbf{Xiao Xu, Nueraili Aierken, Shijian Li$^{\dagger}$} \\[3.5pt]
\textbf{State Key Laboratory of Brain–Machine Intelligence } \\[3.5pt]
\textbf{College of Computer Science and Technology, Zhejiang University, Hangzhou, China} \\[3.5pt]
\textbf{\{boyang30, 22421173, 22351048, 22451116, 3200105334, nureli, shijianli\}@zju.edu.cn}
}

\begin{document}
\maketitle
\begin{abstract}

Intelligent agent systems in real-world agricultural scenarios must handle diverse tasks under multimodal inputs, ranging from lightweight information understanding to complex multi-step execution. However, most existing approaches rely on a unified execution paradigm, which struggles to accommodate large variations in task complexity and incomplete tool availability commonly observed in agricultural environments. To address this challenge, we propose AgriAgent, a two-level agent framework for real-world agriculture. AgriAgent adopts a hierarchical execution strategy based on task complexity: simple tasks are handled through direct reasoning by modality-specific agents, while complex tasks trigger a contract-driven planning mechanism that formulates tasks as capability requirements and performs capability-aware tool orchestration and dynamic tool generation, enabling multi-step and verifiable execution with failure recovery. Experimental results show that AgriAgent achieves higher execution success rates and robustness on complex tasks compared to existing tool-centric agent baselines that rely on unified execution paradigms. All code, data will be released at after our work be accepted to promote reproducible research.

\end{abstract}

\section{Introduction}

\begin{figure}[t]
  \includegraphics[width=\columnwidth]{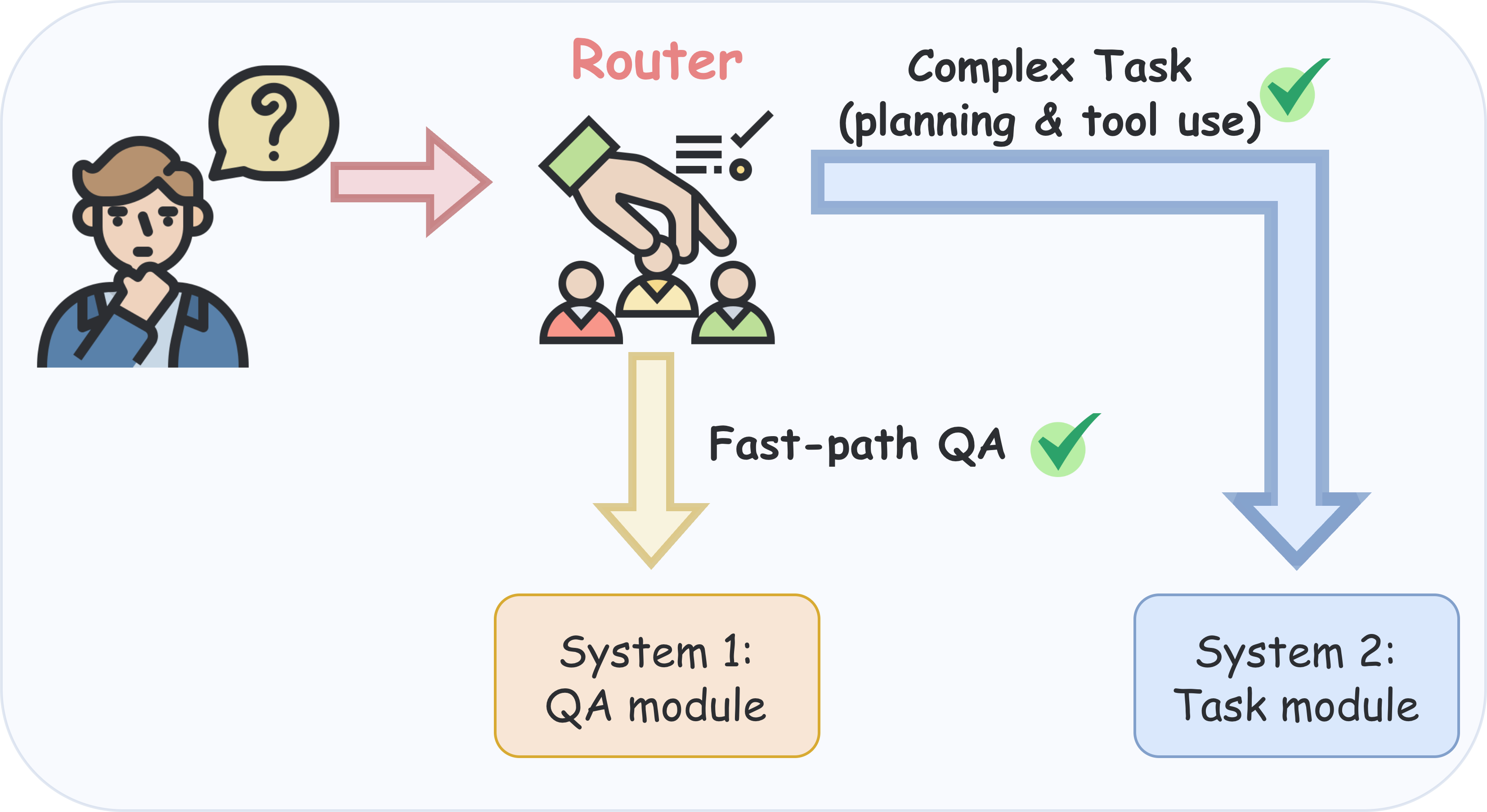}
  \caption{A \textbf{Router} first analyzes the user query and routes it by task complexity.
\textbf{Simple queries} are sent to \textbf{System~1} for fast-path QA without planning or tool use.
\textbf{Complex tasks} are routed to \textbf{System~2} for contract-driven planning and tool execution.}

  \label{fig:experiments}
\end{figure}

\begin{figure*}[t]
  \centering
  \includegraphics[width=\textwidth]{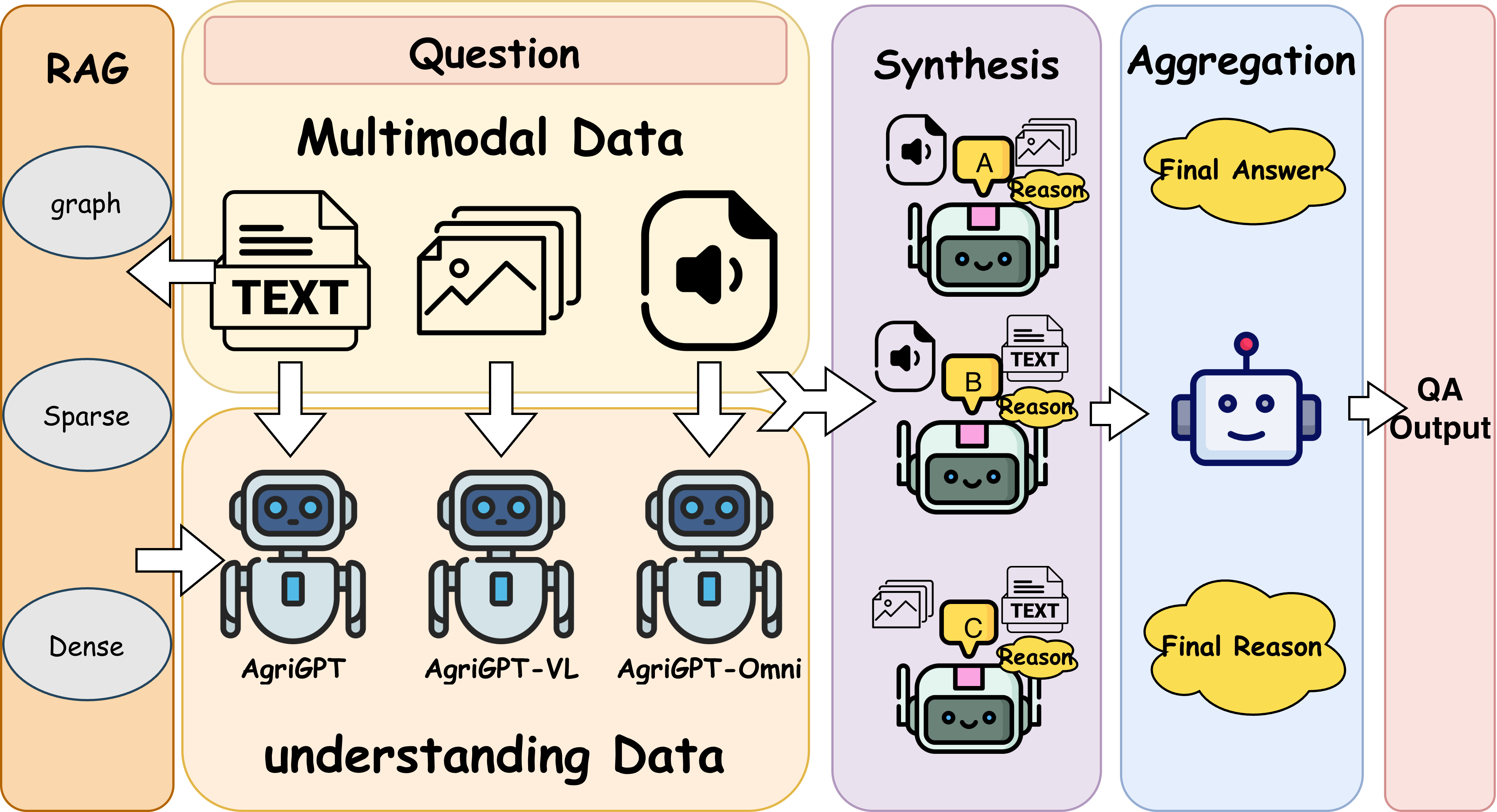}
  \caption{
\textbf{System~1 (Fast Path).}
Multimodal inputs are processed by \textbf{AgriGPT}, \textbf{AgriGPT-VL}, and \textbf{AgriGPT-Omni}, optionally augmented by \textbf{RAG} (dense, sparse, and graph paths).
Independent reasoning results are fused by \textbf{Synthesis} and finalized through \textbf{Aggregation} to produce the QA output.
This path avoids planning and tool orchestration, enabling fast and stable responses for simple agricultural queries.
}

  \label{fig:experiments}
\end{figure*}

Large language model (LLM) based agents have shown strong potential for reasoning and tool-augmented decision making. \citep{yao2022react,schick2023toolformer,nakano2021webgpt,karpas2022mrkl,ahn2022can,patil2024gorilla} However, real-world agricultural tasks pose unique challenges, including long execution chains, multimodal inputs, strong inter-step dependencies, and strict structural constraints. \citep{wang2023voyager,wu2023visual,achiam2023gpt,shen2023hugginggpt,wang2024survey} These characteristics expose limitations of existing agent frameworks in robustness, verifiability, and scalability.\citep{shinn2023reflexion,gou2023critic,madaan2023self,liu2023agentbench,xu2023rewoo}

Most prior agent paradigms rely on implicit language-based planning, where action sequences and tool usage are generated without explicit structural constraints.\citep{yao2022react,yao2023tree,wang2023plan,kojima2022large,wei2022chain} While effective for short or lightweight tasks, such designs often fail in complex agricultural workflows, where missing steps, incorrect dependencies, or inappropriate tool choices lead to cascading execution errors.\citep{shinn2023reflexion,gou2023critic,madaan2023self,liu2023agentbench,wang2022self} As tool ecosystems grow, the challenge further shifts from simply invoking tools to reliably matching task requirements with tool capabilities.\citep{li2023api,qin2023toolllm,patil2024gorilla,schick2023toolformer,shen2023hugginggpt,wang2024survey}

We address these challenges by proposing \textbf{AgriAgent}, an agent framework for real-world agriculture built on \emph{Contract-Driven Planning} and \emph{Capability-Aware Tool Orchestration}.\citep{qin2023toolllm,li2023camel,park2023generative,wang2024survey} AgriAgent adopts a layered design.\citep{kahneman2011thinking} \textbf{System-1} handles structurally simple queries with efficient and consistent understanding, while \textbf{System-2} is responsible for executing complex long-chain tasks through iterative planning, verification, and tool coordination.\citep{shinn2023reflexion,gou2023critic,madaan2023self,liu2023agentbench,xu2023rewoo,wei2022chain,wu2024autogen}

At the core of System-2, we introduce contract-driven planning, which encodes task requirements, input--output constraints, and inter-step dependencies as explicit, verifiable contracts.\citep{meyer1997object,meyer2002design} This transforms planning correctness from an implicit assumption into a checkable property.\citep{meyer1997object,meyer2002design} In addition, we propose a capability-aware tool orchestration mechanism, where a unified ToolHub retrieves and composes tools based on explicit capability descriptions,\citep{li2023api,qin2023toolllm,patil2024gorilla,schick2023toolformer,shen2023hugginggpt,wang2024survey,park2023generative,hong2023metagpt} and a ToolMaker dynamically constructs new tools when required, enabling robust execution in open and evolving environments.\citep{schick2023toolformer,wu2023visual,li2023camel,wu2024autogen}

Our main contributions are summarized as follows:

\begin{itemize}
    \item \textbf{AgriAgent.} 
    We propose \textbf{AgriAgent}, the first layered agent framework in the agricultural domain designed for real-world task execution, which explicitly separates fast-path understanding from complex long-chain execution to support both efficient responses and structured decision-making.

    \item \textbf{Contract-Driven Planning.} 
    We introduce a contract-driven planning mechanism that represents agent execution as explicit and verifiable contracts, improving the controllability, consistency, and interpretability of complex multi-step task planning.

    \item \textbf{Capability-Aware Tool Orchestration.} 
    We propose a capability-aware tool orchestration framework that dynamically aligns task contracts with tool capabilities, enabling adaptive tool retrieval, composition, and on-demand tool construction in open agricultural environments.

    \item \textbf{Extensive Evaluation.} 
    Extensive experiments across diverse agricultural tasks demonstrate that AgriAgent consistently achieves improved robustness and execution reliability compared to strong agent-based baselines, with particularly clear advantages in long-horizon and tool-intensive scenarios.
\end{itemize}

\begin{figure*}[t]
  \centering
  \includegraphics[width=\textwidth]{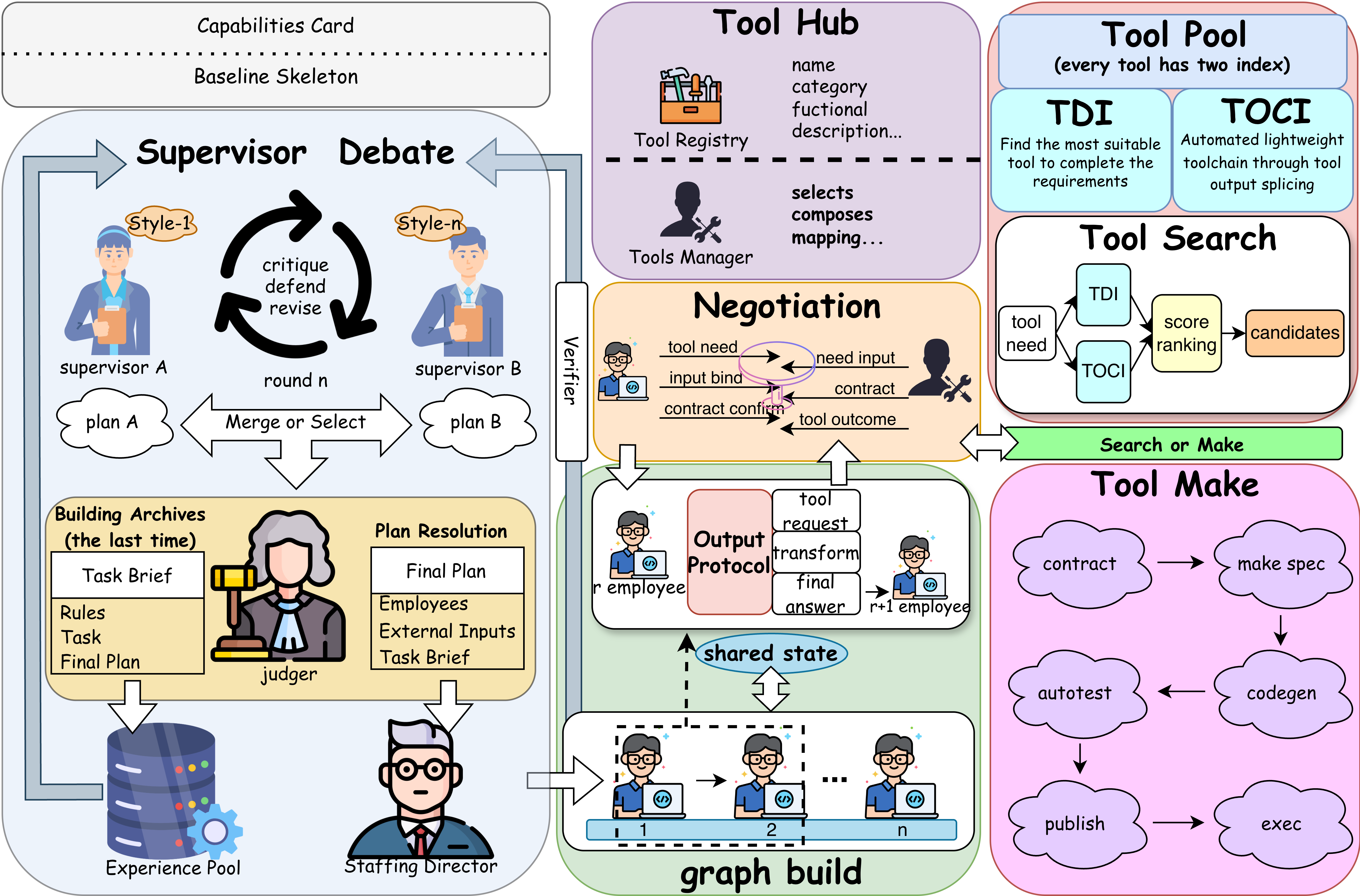}
    \caption{\textbf{System~2 (Complex Task Path).}
    The \textbf{plan/spec} is refined through \textbf{Supervisor Debate}
    (\textbf{critique--defend--revise}) to fix structural errors and missing dependencies.
    Each plan node is then assigned a \textbf{need contract} in the \textbf{Negotiation} stage.
    Based on these contracts, the \textbf{Tool Hub} retrieves and composes tools from the
    \textbf{Tool Pool} via \textbf{TDI/TOCI}, while the \textbf{Tool Maker} generates new tools
    when required. Execution is validated
    to produce a \textbf{verifiable result}.}

  \label{fig:experiments}
\end{figure*}

\begin{figure*}[t]
  \centering
  \includegraphics[width=\textwidth]{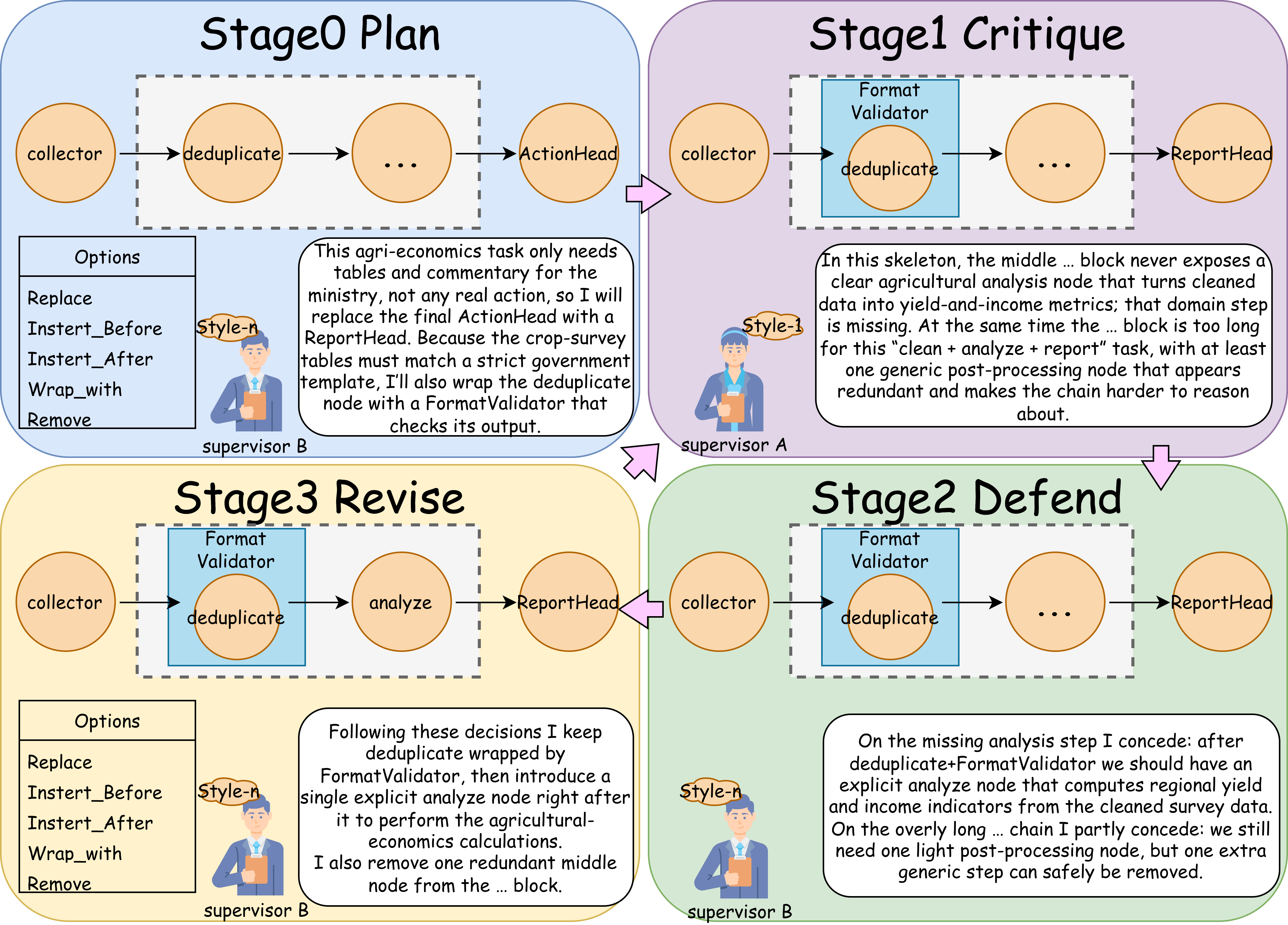}
  \caption{\textbf{Multi-agent roundtable discussion following by critique–defend–revise.}
An initial plan skeleton is generated in \textbf{Stage 0 (Plan)}.
In \textbf{Stage 1 (Critique)}, a supervisor identifies missing, redundant, or invalid nodes.
\textbf{Stage 2 (Defend)} justifies necessary components and resolves disagreements.
\textbf{Stage 3 (Revise)} applies explicit edits (insert, replace, wrap, remove) to produce an executable plan, which is then passed to downstream tool orchestration.}
  \label{fig:experiments}
\end{figure*}

\section{Related Work}

\paragraph{Agents in real-world agriculture.}
\citep{yang2025agrigpt,yang2025agrigpt-VL,yang2025agrigpt-Omni}, while agricultural autonomy has long relied on hierarchical and layered execution for long-term operation \citep{eiffert2022resource,herron2025hierarchical}. 
AgriAgent follows this hierarchical tradition but introduces a router-based two-level execution to avoid over-processing simple requests and brittleness in complex tasks.

\paragraph{Contract-driven planning.}
Hierarchical and anticipatory planning for agricultural robotics has been studied in scheduling, task allocation, and conditional decision frameworks \citep{pal2022agricultural,chen2025multi,khosravi2025optimizing}, and LLM agents further enable multi-step reasoning via deliberative planning \citep{yao2022react,yao2023tree,wang2023plan}. 
AgriAgent differs by enforcing explicit plan specifications and need contracts with debate-based refinement, preventing failures caused by implicit dependencies or missing constraints.

\paragraph{Capability-aware tool orchestration.}
Tool-augmented LLMs are explored through tool learning and benchmarks \citep{schick2023toolformer,patil2024gorilla,qin2023toolllm,xu2023tool,guo2024stabletoolbench,li2023api}, while interactive and dynamic settings emphasize robustness, coordination, and recovery \citep{liu2023agentbench,shridhar2020alfworld,agyeman2025semi,escriba2024digital,yang2025agripath}. 
AgriAgent treats tools as contract-governed capability providers, enabling need-first retrieval, verifiable composition, and reliable execution at scale.

\section{Method}

AgriAgent adopts a two-level agent architecture designed to accommodate the substantial heterogeneity of task complexity encountered in real-world agricultural scenarios. Instead of relying on a single execution paradigm, the system explicitly decomposes task execution into two complementary pathways: \textbf{System~1}, which handles tasks solvable through direct multimodal reasoning, and \textbf{System~2}, which targets complex tasks requiring structured planning, tool interaction, and verifiable execution. Upon receiving a task, AgriAgent first performs unified task understanding and routing, and then dynamically selects the appropriate execution pathway, enabling a balanced trade-off between efficiency, reliability, and extensibility. 

\subsection{Task Definition and Routing}

Given a user instruction $U$, AgriAgent may additionally receive contextual information $C$, including crop type, geographic region, temporal scope, resource constraints, or policy conditions. The system further has access to a registered tool set $T$, external data or knowledge sources $K$, and an environment state $S$ capturing historical execution traces or intermediate states.

The overall objective of AgriAgent is to map these inputs to an executable and verifiable output:
\[
(U, C, T, K, S) \;\rightarrow\; (P, \tau, Y),
\]
where $P$ denotes a structured plan specification, $\tau$ represents tool execution traces, and $Y$ is the final deliverable, such as a natural language response, an analytical report, a structured table, or executable instructions.

At the system entry point, AgriAgent performs task routing based on semantic complexity and execution requirements. Tasks that can be resolved through single-step or weakly dependent reasoning, without explicit external capability invocation or verifiable intermediate outputs, are routed to System~1. Tasks involving multi-step dependencies, multi-source evidence integration, external tool usage, or execution traceability requirements are routed to System~2. By making the decision of whether to plan an explicit system-level operation, AgriAgent avoids the mismatch introduced by a uniform execution paradigm.

\subsection{System-1: Fast-path Multimodal Question Answering}
\label{subsec:system1}

In real agricultural applications, many user queries do not require explicit action planning but involve heterogeneous inputs such as text, field images, and spoken instructions.
Relying on a single omnimodal model in such settings often leads to unstable performance due to modality imbalance, knowledge inconsistency, or amplified noise.
System-1 addresses this challenge through a system-level design that combines modality-specialized models with unified retrieval augmentation, achieving efficient and reliable multimodal understanding.

\paragraph{Modality-Aware Processing and Model Collaboration.}
System-1 processes three input modalities: text, images, and audio, which differ substantially in structure and noise characteristics.
Instead of early fusion into a single representation, each modality is handled by a specialized module.
\textbf{AgriGPT} focuses on text-based semantic parsing and domain knowledge,
\textbf{AgriGPT-VL} grounds visual observations in agricultural concepts,
and \textbf{AgriGPT-Omni} integrates intermediate representations to form a unified semantic interpretation.
This collaborative design improves robustness across diverse input conditions and avoids the trade-offs commonly observed in single omnimodal models.

\paragraph{Unified Retrieval-Augmented Reasoning.}
To ensure factual grounding, System-1 incorporates a unified Retrieval-Augmented Generation (RAG) mechanism shared across all modules.
Three complementary retrieval paths are employed: dense retrieval for semantic relevance, sparse retrieval for precise keyword matching, and graph-based retrieval for structured or multi-hop relations.
All retrieved evidence is consolidated into a shared context, enabling consistent knowledge access and alignment across modalities.

\paragraph{Collaborative Answer Synthesis.}
Based on modality-specific interpretations and retrieved evidence, System-1 adopts a collaborative chain-based reasoning strategy.
Each module produces intermediate semantic judgments grounded in its respective modality, which are subsequently fused by AgriGPT-Omni to generate the final answer.

\begin{table*}[t]
\centering
\small

\begin{adjustbox}{width=\textwidth}
\begin{tabular}{llcccccccccccc}
\toprule
\multirow{2}{*}{Model} & \multirow{2}{*}{Method} &
\multicolumn{4}{c}{\textbf{Programmatic}} &
\multicolumn{4}{c}{\textbf{LLM}} &
\multicolumn{4}{c}{\textbf{LLM (Non-Zero)}} \\
\cmidrule(lr){3-6} \cmidrule(lr){7-10} \cmidrule(lr){11-14}
& &
Pres. & Rule & Evidence & Norm. &
Task & Process & Rule & Evidence &
Task & Process & Rule & Evidence \\
& &
Cov. & Cit. & Pres. & Via-Bus &
Fulfill. & Fulfill. & Usage & Usage &
Fulfill. & Fulfill. & Usage & Usage \\
\midrule

\multirow{4}{*}{Qwen2.5-7B}
& react & 0.121 & 0.000 & 0.000 & 0.350 & 0.070 & 0.667 & 0.022 & 0.047 & 0.589 & 0.667 & 0.297 & 0.432 \\
& plan-and-execute & 0.593 & 0.013 & 0.100 & 0.420 & 0.396 & 0.693 & 0.180 & 0.289 & 0.670 & 0.693 & 0.341 & 0.486 \\
& ours-w/o-fb & 0.559 & 0.373 & 0.392 & 0.543 & 0.375 & \textcolor{red}{0.705} & 0.339 & 0.312 & 0.504 & \textcolor{red}{0.705} & 0.478 & 0.418 \\
& ours & \textcolor{red}{0.944} & \textcolor{red}{0.719} & \textcolor{red}{0.757} & \textcolor{red}{0.555}
& \textcolor{red}{0.719} & 0.702 & \textcolor{red}{0.619} & \textcolor{red}{0.605}
& \textcolor{red}{0.763} & 0.702 & \textcolor{red}{0.674} & \textcolor{red}{0.641} \\

\midrule

\multirow{4}{*}{Qwen3-8B}
& react & 0.437 & 0.003 & 0.003 & 0.650 & 0.215 & 0.738 & 0.123 & 0.196 & 0.496 & 0.738 & 0.315 & 0.448 \\
& plan-and-execute & 0.433 & 0.048 & 0.067 & 0.610 & 0.264 & 0.761 & 0.181 & 0.235 & 0.612 & 0.761 & 0.451 & 0.540 \\
& ours-w/o-fb & 0.967 & \textcolor{red}{0.662} & \textcolor{red}{0.664} & 0.786
& 0.669 & \textcolor{red}{0.771} & 0.628 & \textcolor{red}{0.580}
& 0.693 & \textcolor{red}{0.771} & \textcolor{red}{0.690} & \textcolor{red}{0.600} \\
& ours & \textcolor{red}{0.987} & 0.497 & 0.505 & \textcolor{red}{0.786}
& \textcolor{red}{0.689} & 0.768 & \textcolor{red}{0.661} & 0.576
& \textcolor{red}{0.703} & 0.768 & 0.672 & 0.589 \\

\midrule

\multirow{4}{*}{LLaMA3-8B}
& react & 0.063 & 0.003 & 0.000 & 0.010 & 0.034 & 0.424 & 0.017 & 0.030 & 0.522 & 0.424 & 0.286 & 0.453 \\
& plan-and-execute & 0.004 & 0.000 & 0.000 & 0.029 & 0.002 & 0.416 & 0.001 & 0.002 & 0.400 & 0.416 & 0.200 & 0.400 \\
& ours-w/o-fb & 0.495 & 0.319 & 0.393 & 0.315 & 0.218 & 0.746 & 0.218 & 0.198 & 0.441 & 0.746 & 0.452 & 0.396 \\
& ours & \textcolor{red}{0.909} & \textcolor{red}{0.673} & \textcolor{red}{0.766} & \textcolor{red}{0.358}
& \textcolor{red}{0.555} & \textcolor{red}{0.748} & \textcolor{red}{0.473} & \textcolor{red}{0.479}
& \textcolor{red}{0.635} & \textcolor{red}{0.748} & \textcolor{red}{0.557} & \textcolor{red}{0.536} \\

\midrule

\multirow{4}{*}{AgriGPT-8B}
& react & 0.538 & 0.008 & 0.026 & 0.620 & 0.267 & 0.717 & 0.142 & 0.226 & 0.505 & 0.717 & 0.326 & 0.423 \\
& plan-and-execute & 0.480 & 0.063 & 0.088 & 0.650 & 0.290 & 0.775 & 0.209 & 0.270 & 0.630 & 0.775 & 0.455 & 0.558 \\
& ours-w/o-fb & 0.919 & \textcolor{red}{0.628} & \textcolor{red}{0.638} & \textcolor{red}{0.722}
& 0.688 & 0.776 & \textcolor{red}{0.639} & 0.631
& \textcolor{red}{0.751} & 0.776 & \textcolor{red}{0.729} & \textcolor{red}{0.688} \\
& ours & \textcolor{red}{0.981} & 0.572 & 0.586 & 0.700
& \textcolor{red}{0.729} & \textcolor{red}{0.790} & 0.622 & \textcolor{red}{0.639}
& 0.745 & \textcolor{red}{0.790} & 0.669 & 0.655 \\

\midrule

\multirow{4}{*}{Qwen2.5-14B}
& react & 0.624 & 0.092 & 0.043 & 0.460 & 0.431 & 0.746 & 0.203 & 0.317 & 0.693 & 0.746 & 0.378 & 0.513 \\
& plan-and-execute & 0.567 & 0.031 & 0.084 & 0.320 & 0.366 & 0.769 & 0.194 & 0.303 & 0.649 & 0.769 & 0.375 & 0.534 \\
& ours-w/o-fb & 0.691 & 0.510 & 0.531 & 0.489 & 0.535 & 0.780 & 0.485 & 0.461 & 0.776 & 0.780 & 0.724 & 0.668 \\
& ours & \textcolor{red}{0.969} & \textcolor{red}{0.741} & \textcolor{red}{0.809} & \textcolor{red}{0.500}
& \textcolor{red}{0.775} & \textcolor{red}{0.786} & \textcolor{red}{0.675} & \textcolor{red}{0.679}
& \textcolor{red}{0.803} & \textcolor{red}{0.786} & \textcolor{red}{0.729} & \textcolor{red}{0.703} \\

\midrule

\multirow{4}{*}{Qwen3-32B}
& react & 0.178 & 0.007 & 0.013 & 0.600 & 0.079 & 0.700 & 0.046 & 0.079 & 0.475 & 0.700 & 0.301 & 0.446 \\
& plan-and-execute & 0.494 & 0.085 & 0.112 & 0.610 & 0.354 & 0.781 & 0.244 & 0.303 & 0.736 & 0.781 & 0.538 & 0.633 \\
& ours-w/o-fb & 0.966 & 0.741 & 0.690 & 0.736 & 0.752 & 0.788 & 0.727 & 0.684 & 0.777 & 0.788 & 0.770 & 0.706 \\
& ours & \textcolor{red}{1.000} & \textcolor{red}{0.786} & \textcolor{red}{0.731} & \textcolor{red}{0.783}
& \textcolor{red}{0.776} & \textcolor{red}{0.796} & \textcolor{red}{0.767} & \textcolor{red}{0.709}
& \textcolor{red}{0.785} & \textcolor{red}{0.796} & \textcolor{red}{0.784} & \textcolor{red}{0.713} \\

\midrule

\multirow{4}{*}{Qwen3-235B}
& react & - & - & - & - & - & - & - & - & - & - & - & - \\
& plan-and-execute & - & - & - & - & - & - & - & - & - & - & - & - \\
& ours-w/o-fb & - & - & - & - & - & - & - & - & - & - & - & - \\
& ours & \textcolor{red}{1.000} & \textcolor{red}{1.000} & \textcolor{red}{1.000} & - & \textcolor{red}{0.872} & - & \textcolor{red}{0.819} & \textcolor{red}{0.798} & \textcolor{red}{0.872} & - & \textcolor{red}{0.819} & \textcolor{red}{0.798} \\

\midrule

\multirow{4}{*}{DeepSeek-v3.1}
& react & - & - & - & - & - & - & - & - & - & - & - & - \\
& plan-and-execute & - & - & - & - & - & - & - & - & - & - & - & - \\
& ours-w/o-fb & - & - & - & - & - & - & - & - & - & - & - & - \\
& ours & \textcolor{red}{0.950} & \textcolor{red}{0.947} & \textcolor{red}{0.950} & - & \textcolor{red}{0.809} & - & \textcolor{red}{0.738} & \textcolor{red}{0.715} & \textcolor{red}{0.851} & - & \textcolor{red}{0.802} & \textcolor{red}{0.753} \\

\bottomrule
\end{tabular}
\end{adjustbox}

\caption{Overall comparison of different models and methods under three complementary evaluation regimes, including programmatic metrics and LLM-based metrics, validating the effectiveness of our overall strategy framework. ``--'' indicates settings where process-level metrics are not applicable, as the model directly produces final outputs without executing intermediate planning or tool-based steps.}

\label{tab:main_results}

\end{table*}

\begin{table*}[t]
\centering
\small
\setlength{\tabcolsep}{2.5pt}
\renewcommand{\arraystretch}{1.05}

\begin{adjustbox}{width=\textwidth}
\begin{tabular}{ll ccc ccc ccc ccc ccc ccc}
\toprule
\multirow{3}{*}{Model} & \multirow{3}{*}{Method}
& \multicolumn{6}{c}{\textbf{24 Tools}}
& \multicolumn{6}{c}{\textbf{48 Tools}}
& \multicolumn{6}{c}{\textbf{Unbounded Tools (506)}} \\
\cmidrule(lr){3-8}\cmidrule(lr){9-14}\cmidrule(lr){15-20}
& 
& \multicolumn{3}{c}{single} & \multicolumn{3}{c}{chain}
& \multicolumn{3}{c}{single} & \multicolumn{3}{c}{chain}
& \multicolumn{3}{c}{single} & \multicolumn{3}{c}{chain} \\
\cmidrule(lr){3-5}\cmidrule(lr){6-8}
\cmidrule(lr){9-11}\cmidrule(lr){12-14}
\cmidrule(lr){15-17}\cmidrule(lr){18-20}
& 
& Hit@1 & Hit@3 & Hit@5 & Hit@1 & Hit@3 & Hit@5
& Hit@1 & Hit@3 & Hit@5 & Hit@1 & Hit@3 & Hit@5
& Hit@1 & Hit@3 & Hit@5 & Hit@1 & Hit@3 & Hit@5 \\
\midrule

\multirow{3}{*}{Qwen2.5-7B}
& top-$k$-prompt
& 0.896 & 0.896 & 0.896 & 0.500 & 0.556 & 0.556
& 0.896 & 0.896 & 0.896 & 0.417 & 0.417 & 0.417
& 0.233 & 0.233 & 0.233 & 0.381 & 0.381 & 0.381 \\
& all-in-prompt
& 0.208 & 0.208 & 0.208 & 0.389 & 0.389 & 0.389
& 0.031 & 0.031 & 0.031 & 0.028 & 0.028 & 0.028
& -- & -- & -- & -- & -- & -- \\
& toolhub
& \textcolor{red}{0.896} & \textcolor{red}{1.000} & \textcolor{red}{1.000}
& \textcolor{red}{0.940} & \textcolor{red}{0.940} & \textcolor{red}{0.940}
& \textcolor{red}{0.865} & \textcolor{red}{1.000} & \textcolor{red}{1.000}
& \textcolor{red}{0.833} & \textcolor{red}{1.000} & \textcolor{red}{1.000}
& \textcolor{red}{0.907} & \textcolor{red}{0.907} & \textcolor{red}{0.907}
& \textcolor{red}{0.766} & \textcolor{red}{0.766} & \textcolor{red}{0.766} \\
\midrule

\multirow{3}{*}{Qwen3-8B}
& top-$k$-prompt
& 1.000 & 1.000 & 1.000 & 0.778 & 0.778 & 0.778
& 1.000 & 1.000 & 1.000 & 0.778 & 0.778 & 0.778
& 0.240 & 0.240 & 0.240 & 0.103 & 0.103 & 0.103 \\
& all-in-prompt
& 0.917 & 0.917 & 0.917 & 0.611 & 0.667 & 0.667
& 0.906 & 0.906 & 0.906 & 0.528 & 0.528 & 0.528
& -- & -- & -- & -- & -- & -- \\
& toolhub
& \textcolor{red}{0.9375} & \textcolor{red}{1.000} & \textcolor{red}{1.000}
& \textcolor{red}{1.000} & \textcolor{red}{1.000} & \textcolor{red}{1.000}
& \textcolor{red}{0.927} & \textcolor{red}{1.000} & \textcolor{red}{1.000}
& \textcolor{red}{0.917} & \textcolor{red}{1.000} & \textcolor{red}{1.000}
& \textcolor{red}{0.900} & \textcolor{red}{0.907} & \textcolor{red}{0.907}
& \textcolor{red}{0.754} & \textcolor{red}{0.754} & \textcolor{red}{0.754} \\
\midrule

\multirow{2}{*}{LLaMA3-8B}
& top-$k$-prompt
& 0.021 & 0.021 & 0.021 & 0.444 & 0.500 & 0.500
& 0.073 & 0.073 & 0.073 & 0.222 & 0.222 & 0.222
& 0.107 & 0.120 & 0.120 & 0.036 & 0.040 & 0.040 \\
& toolhub
& \textcolor{red}{0.9375} & \textcolor{red}{0.958} & \textcolor{red}{0.958}
& \textcolor{red}{1.000} & \textcolor{red}{1.000} & \textcolor{red}{1.000}
& \textcolor{red}{0.875} & \textcolor{red}{0.917} & \textcolor{red}{0.917}
& \textcolor{red}{0.833} & \textcolor{red}{1.000} & \textcolor{red}{1.000}
& \textcolor{red}{0.913} & \textcolor{red}{0.913} & \textcolor{red}{0.913}
& \textcolor{red}{0.813} & \textcolor{red}{0.813} & \textcolor{red}{0.813} \\
\midrule

\multirow{3}{*}{Agrigpt-8B}
& top-$k$-prompt
& 1.000 & 1.000 & 1.000 & 0.889 & 0.889 & 0.889
& 1.000 & 1.000 & 1.000 & 0.833 & 0.833 & 0.833
& 0.233 & 0.233 & 0.233 & 0.1032 & 0.1032 & 0.1032 \\
& all-in-prompt
& 0.917 & 0.917 & 0.917 & 0.889 & 0.889 & 0.889
& 0.906 & 0.906 & 0.906 & 0.583 & 0.583 & 0.583
& -- & -- & -- & -- & -- & -- \\
& toolhub
& \textcolor{red}{0.958} & \textcolor{red}{1.000} & \textcolor{red}{1.000}
& \textcolor{red}{1.000} & \textcolor{red}{1.000} & \textcolor{red}{1.000}
& \textcolor{red}{0.9375} & \textcolor{red}{1.000} & \textcolor{red}{1.000}
& \textcolor{red}{0.833} & \textcolor{red}{1.000} & \textcolor{red}{1.000}
& \textcolor{red}{0.907} & \textcolor{red}{0.907} & \textcolor{red}{0.907}
& \textcolor{red}{0.754} & \textcolor{red}{0.754} & \textcolor{red}{0.754} \\
\midrule

\multirow{3}{*}{Qwen2.5-14B}
& top-$k$-prompt
& 0.708 & 0.792 & 0.792 & 0.944 & 1.000 & 1.000
& 0.771 & 0.896 & 0.896 & 0.889 & 0.889 & 0.889
& 0.353 & 0.353 & 0.353 & 0.698 & 0.702 & 0.702 \\
& all-in-prompt
& 0.146 & 0.250 & 0.250 & 0.667 & 0.667 & 0.667
& 0.104 & 0.156 & 0.156 & 0.694 & 0.694 & 0.694
& -- & -- & -- & -- & -- & -- \\
& toolhub
& \textcolor{red}{0.958} & \textcolor{red}{1.000} & \textcolor{red}{1.000}
& \textcolor{red}{1.000} & \textcolor{red}{1.000} & \textcolor{red}{1.000}
& \textcolor{red}{0.917} & \textcolor{red}{1.000} & \textcolor{red}{1.000}
& \textcolor{red}{0.833} & \textcolor{red}{1.000} & \textcolor{red}{1.000}
& \textcolor{red}{0.907} & \textcolor{red}{0.907} & \textcolor{red}{0.907}
& \textcolor{red}{0.770} & \textcolor{red}{0.770} & \textcolor{red}{0.770} \\
\midrule

\multirow{3}{*}{Qwen3-32B}
& top-$k$-prompt
& 1.000 & 1.000 & 1.000 & 1.000 & 1.000 & 1.000
& 0.948 & 0.948 & 0.948 & 0.833 & 0.861 & 0.861
& 0.320 & 0.320 & 0.320 & 0.571 & 0.571 & 0.571 \\
& all-in-prompt
& 0.458 & 0.479 & 0.479 & 0.722 & 0.722 & 0.722
& 0.490 & 0.490 & 0.510 & 0.583 & 0.583 & 0.583
& -- & -- & -- & -- & -- & -- \\
& toolhub
& \textcolor{red}{0.958} & \textcolor{red}{1.000} & \textcolor{red}{1.000}
& \textcolor{red}{0.889} & \textcolor{red}{1.000} & \textcolor{red}{1.000}
& \textcolor{red}{0.9375} & \textcolor{red}{1.000} & \textcolor{red}{1.000}
& \textcolor{red}{0.833} & \textcolor{red}{1.000} & \textcolor{red}{1.000}
& \textcolor{red}{0.907} & \textcolor{red}{0.907} & \textcolor{red}{0.907}
& \textcolor{red}{0.766} & \textcolor{red}{0.766} & \textcolor{red}{0.766} \\
\bottomrule
\end{tabular}
\end{adjustbox}
\caption{\textbf{Tool selection performance under different tool scales and invocation settings.}
We compare ToolHub with prompt-based baselines under single-step and multi-step (chain) tool invocation.}
\label{tab:toolhub}
\end{table*}

\subsection{System~2: Contract-Driven Structured Planning and Execution}

System~2 addresses complex tasks that require multi-step decision-making, integration of heterogeneous evidence, interaction with external tools, and explicit control over execution and verification. Its core principle is to decouple task requirements from concrete tool implementations through structured representations and capability contracts. 

\subsubsection{Structured Plan Specification}

In System~2, each task is first represented as a structured plan specification in the form of a \emph{Directed Acyclic Graph (DAG)}:
\[
P = (V, E),
\]
where each node $v_i \in V$ corresponds to an atomic execution unit, and each edge $e_{ij} \in E$ denotes a dependency between nodes. Each node is formalized as:
\[
v_i = \langle g_i, I_i, O_i, \mathcal{C}_i, \mathcal{E}_i \rangle,
\]
where $g_i$ denotes the local goal, $I_i$ and $O_i$ denote required inputs and expected outputs, $\mathcal{C}_i$ represents execution constraints (e.g., format, scope, or policy requirements), and $\mathcal{E}_i$ specifies evidence or verification requirements.

This explicit representation transforms implicit multi-step reasoning into a manipulable and inspectable structure, enabling systematic planning refinement, execution control, and error localization.

\subsubsection{Need Contracts}

To bridge structured plans and executable capabilities, System~2 associates each plan node with a \emph{need contract}. A need contract abstracts execution requirements at the capability level rather than binding them to specific tools:
\[
c_i = \langle \textit{cap}_i,\; \mathcal{S}^{in}_i,\; \mathcal{S}^{out}_i,\; pre_i,\; con_i,\; q_i \rangle,
\]
where $\textit{cap}_i$ specifies the required capability, $\mathcal{S}^{in}_i$ and $\mathcal{S}^{out}_i$ define input and output schemas, $pre_i$ denotes preconditions, $con_i$ denotes execution constraints, and $q_i$ specifies quality or success criteria.

By explicitly separating \emph{what capability is required} from \emph{how it is implemented}, need contracts enable flexible tool selection, substitution, and extension under changing tool availability.

\subsubsection{Debate-Based Plan Generation and Refinement}

Initial plan specifications are generated independently by multiple supervisor agents employing diverse reasoning preferences. These candidate plans are subsequently refined through an iterative debate process consisting of \emph{critique}, \emph{defense}, and \emph{revision} phases.

During critique, supervisors identify missing steps, redundant nodes, inconsistent dependencies, or implicit assumptions. In the defense phase, existing design choices are evaluated against constraints and feasibility considerations. The revision phase applies explicit structural edits to the DAG, including node insertion, replacement, wrapping, or removal. After multiple iterations, the system selects or merges plans to obtain a structurally consistent and executable specification.

This process surfaces planning assumptions early and reduces the propagation of structural errors into execution.

\paragraph{Employee--ToolHub Negotiation.}
In AgriAgent, tools are not invoked directly by the agent.
Instead, the Employee negotiates with the ToolHub by declaring an explicit \emph{tool need}, receiving candidate input requirements, and confirming a contract that binds all required inputs and capability constraints.
Only after contract confirmation is the tool executed and its outcome returned to the Employee for subsequent steps.

\subsubsection{Tool Hub and Dual-Protocol Tool Matching}

System~2 manages tools through a centralized \emph{Tool Hub}, where each tool is registered via a tool card specifying its capabilities, input/output schemas, preconditions, constraints, reliability signals, and provenance.

Tool matching is governed by two complementary protocols:
\begin{itemize}
  \item \textbf{Tool Description Index (TDI):} retrieves candidate tools by matching required capabilities in need contracts against tool metadata embeddings, producing a capability-consistent candidate set.
  \item \textbf{Tool Output Composition Index (TOCI):} verifies schema compatibility among candidate tools by aligning output--input specifications, enabling lightweight composition into executable tool chains. Incompatible or partially satisfied schemas are filtered out before execution.

\end{itemize}

By decoupling capability matching from schema-level composition, TDI and TOCI together enable capability-aware orchestration without reliance on fixed tool names or hard-coded pipelines. 
This design supports scalable tool selection under large and evolving tool inventories, while allowing third-party tools to be integrated with minimal assumptions.

\subsubsection{Dynamic Tool Generation}

When no existing tool satisfies a given need contract, or when candidate tools repeatedly fail during execution or verification, System~2 activates a \emph{Tool Maker} module. Tool Maker derives a tool specification directly from the unmet contract, generates an implementation, validates it through automated testing, and registers it back into the Tool Hub.

This mechanism converts capability gaps into verifiable executable modules, enabling the system to adapt to incomplete or evolving tool ecosystems rather than relying on speculative reasoning.

\subsubsection{Execution, Verification, and Evidence Aggregation}

The finalized plan is executed following the DAG dependencies. Each execution step enforces schema validation and contract consistency checks, while recording tool invocation traces and intermediate outputs. Evidence is aggregated along dependency paths to produce a structured final output accompanied by provenance information.

By tightly coupling execution with verification and evidence aggregation, System~2 delivers results that are not only correct but also auditable, reproducible, and suitable for iterative refinement.

\section{Results}

\subsection{Metrics}
\label{sec:metrics}

We evaluate AgriAgent using both programmatic metrics and LLM-based metrics,
targeting complementary aspects of execution correctness and semantic task fulfillment.

\paragraph{Programmatic metrics.}
Programmatic metrics are computed deterministically from the final execution results.
\emph{Presence Coverage} measures whether all required fields specified in the need contracts
are instantiated in the final output.
\emph{Rule Citation} checks whether required business or operational rules are explicitly referenced.
\emph{Evidence Presence} verifies whether claims are supported by retrieved or tool-generated evidence.
\emph{Normalization via Business Rules} evaluates whether outputs conform to predefined schema and normalization constraints.

\paragraph{LLM-based metrics.}
LLM-based metrics assess semantic task completion and reasoning quality using a fixed judge model.
These metrics serve as approximate proxies and do not replace human evaluation.

\paragraph{Non-Zero evaluation.}
For methods that may produce empty or degenerate outputs, we additionally report \textbf{LLM (Non-Zero)},
which computes the same LLM-based metrics only on non-empty outputs to diagnose robustness independently
of execution failures.

\paragraph{Evaluation Datasets.}
System-1 is evaluated on the AgriGPT-Omni dataset \citep{yang2025agrigpt-Omni} for multimodal agricultural question answering, while System-2 is evaluated on a separately constructed dataset of 1,000 complex agricultural tasks.

\begin{table}[t]
\centering
\small

\begin{tabular}{l l l c}
\toprule
\textbf{Model Family} & \textbf{Composition} & \textbf{Reasoning} & \textbf{Score} \\
\midrule
\multirow{3}{*}{AgriGPT (7B)}
& Omni & CoT & 0.80 \\
& Omni & ToT & 0.69 \\
& Text + VL + Omni & \textbf{Ours} & \textcolor{red}{0.87} \\
\midrule
\multirow{3}{*}{Qwen2.5 (7B)}
& Omni & CoT & 0.79 \\
& Omni & ToT & 0.60 \\
& Text + VL + Omni & \textbf{Ours} & \textcolor{red}{0.83} \\
\midrule
\multirow{3}{*}{Qwen3 (30B)}
& Omni & CoT & 0.88 \\
& Omni & ToT & 0.93 \\
& Text + VL + Omni & \textbf{Ours} & \textcolor{red}{0.94} \\
\bottomrule
\end{tabular}
\caption{Comparison of single-model reasoning and multi-modal fusion. Score denotes averaged task completion accuracy on multimodal QA tasks.}

\label{tab:multimodal_reasoning}
\end{table}

\subsection{System-1 Results}
\label{subsec:system1_results}

We evaluate \textbf{System-1} on fast-path multimodal question answering in agricultural scenarios,
comparing it with single omni-model reasoning strategies based on Chain-of-Thought (CoT)~\citep{wei2022chain} and Tree-of-Thought (ToT)~\citep{yao2023tree}.

As shown in Table~\ref{tab:multimodal_reasoning}, experiments cover multiple model scales,
including AgriGPT and Qwen omni models.
For each setting, we compare single-model CoT/ToT reasoning with \textbf{System-1 (Ours)},
which integrates text, vision, and omni-modal models via coordinated reasoning.
All methods share the same RAG mechanism to ensure fair comparison.

Results show that \textbf{System-1 consistently outperforms CoT and ToT across all model scales},
with larger gains on smaller and medium-sized models.
This suggests that improvements stem from modality-specialized collaboration and shared retrieval grounding,
rather than increased reasoning depth within a single model.

\begin{table}[t]
\centering
\small

\begin{tabular}{lcccc}
\toprule

\textbf{Component} & \textbf{Attempts} & \textbf{Succeeded} & \textbf{Failed} & \textbf{Rate} \\
\midrule
ToolMaker & 392 & 380 & 12 & 0.9694 \\
\bottomrule
\end{tabular}
\caption{ToolMaker execution statistics in System-2.} 
\label{tab:toolmaker}
\end{table}

\subsection{System-2 Results}

System-2 targets complex multi-step agricultural tasks.
We evaluate it on: (i) contract-driven planning effectiveness, (ii) ToolHub tool selection under scaling, and (iii) ToolMaker reliability.

\paragraph{Contract-Driven Planning.}
Table~\ref{tab:main_results} shows our contract-driven planning consistently outperforms react and plan-and-execute across model families.
Gains hold for both programmatic metrics (rule citation, evidence presence, business-rule normalization) and LLM-based metrics,
indicating that executable contracts reduce structurally invalid yet fluent plans.
Consistent improvements on general-purpose and domain models suggest the benefit comes from the framework rather than model-specific capability.

\paragraph{ToolHub for Capability-Aware Orchestration.}
Table~\ref{tab:toolhub} compares tool selection under different tool scales and single/chain invocation.
Prompt baselines (all-in-prompt, top-$k$) degrade sharply as the tool pool grows,
especially in the 506-tool unbounded setting evolved from \emph{ToolHop}~\cite{ye2025toolhop}.
ToolHub keeps stable Hit@1/3/5 across scales and models, with larger gains in chain invocation,
where capability modeling and contract constraints matter for composing dependent tools.

\paragraph{ToolMaker Reliability.}
Table~\ref{tab:toolmaker} reports 380/392 successful tool generations (96.94\%).
Failures are mainly due to external environment constraints rather than planning/contract violations,
showing dynamic tool construction is feasible and reliable when guided by explicit contracts and capability specs.

\section{Conclusion}

We introduce AgriAgent, a layered agent system for real-world agriculture.
System-1 handles fast-path multimodal question answering,
while System-2 executes complex tasks via contract-driven planning and capability-aware tool orchestration.
By formalizing task requirements as verifiable contracts, AgriAgent enables reliable tool selection, composition, and dynamic tool generation in high-constraint environments.

\section*{Limitations}   

\begin{itemize}
  \item \textbf{Computational and interaction overhead in long-horizon tasks.} 
  For complex long-horizon tasks, System-2 requires multiple rounds of planning, contract verification, and tool invocation, which introduce additional computational cost and interaction latency. While this design improves execution reliability and structural correctness, it may be less suitable for scenarios with strict real-time requirements, where further trade-offs between efficiency and robustness are necessary.

  \item \textbf{Dependence on structured output contracts and schema adherence.} 
  Contract-driven planning relies on models’ ability to consistently follow predefined output schemas and key structures. When using smaller or less capable models, deviations from the expected formats may affect contract parsing and downstream execution. Improving robustness to format variations remains an important direction for future work.

  \item \textbf{Scope of tool construction in ToolMaker.} 
  ToolMaker currently focuses on general-purpose tool construction, including requirement confirmation, tool generation, validation, and registration. While effective for many practical scenarios, it is not explicitly optimized for synthesizing highly complex or domain-specific tools, which may require more advanced generation and verification pipelines.

  \item \textbf{We do not report detailed latency or token cost analysis, which is an important direction for future work.} 

\end{itemize}

\bibliography{custom}

@inproceedings{yao2022react,
  title={React: Synergizing reasoning and acting in language models},
  author={Yao, Shunyu and Zhao, Jeffrey and Yu, Dian and Du, Nan and Shafran, Izhak and Narasimhan, Karthik R and Cao, Yuan},
  booktitle={The eleventh international conference on learning representations},
  year={2022}
}

@article{schick2023toolformer,
  title={Toolformer: Language models can teach themselves to use tools},
  author={Schick, Timo and Dwivedi-Yu, Jane and Dess{\`\i}, Roberto and Raileanu, Roberta and Lomeli, Maria and Hambro, Eric and Zettlemoyer, Luke and Cancedda, Nicola and Scialom, Thomas},
  journal={Advances in Neural Information Processing Systems},
  volume={36},
  pages={68539--68551},
  year={2023}
}

@article{nakano2021webgpt,
  title={Webgpt: Browser-assisted question-answering with human feedback},
  author={Nakano, Reiichiro and Hilton, Jacob and Balaji, Suchir and Wu, Jeff and Ouyang, Long and Kim, Christina and Hesse, Christopher and Jain, Shantanu and Kosaraju, Vineet and Saunders, William and others},
  journal={arXiv preprint arXiv:2112.09332},
  year={2021}
}

@article{karpas2022mrkl,
  title={MRKL Systems: A modular, neuro-symbolic architecture that combines large language models, external knowledge sources and discrete reasoning},
  author={Karpas, Ehud and Abend, Omri and Belinkov, Yonatan and Lenz, Barak and Lieber, Opher and Ratner, Nir and Shoham, Yoav and Bata, Hofit and Levine, Yoav and Leyton-Brown, Kevin and others},
  journal={arXiv preprint arXiv:2205.00445},
  year={2022}
}

@article{ahn2022can,
  title={Do as i can, not as i say: Grounding language in robotic affordances},
  author={Ahn, Michael and Brohan, Anthony and Brown, Noah and Chebotar, Yevgen and Cortes, Omar and David, Byron and Finn, Chelsea and Fu, Chuyuan and Gopalakrishnan, Keerthana and Hausman, Karol and others},
  journal={arXiv preprint arXiv:2204.01691},
  year={2022}
}

@article{patil2024gorilla,
  title={Gorilla: Large language model connected with massive apis},
  author={Patil, Shishir G and Zhang, Tianjun and Wang, Xin and Gonzalez, Joseph E},
  journal={Advances in Neural Information Processing Systems},
  volume={37},
  pages={126544--126565},
  year={2024}
}

@article{wang2023voyager,
  title={Voyager: An open-ended embodied agent with large language models},
  author={Wang, Guanzhi and Xie, Yuqi and Jiang, Yunfan and Mandlekar, Ajay and Xiao, Chaowei and Zhu, Yuke and Fan, Linxi and Anandkumar, Anima},
  journal={arXiv preprint arXiv:2305.16291},
  year={2023}
}

@article{wu2023visual,
  title={Visual chatgpt: Talking, drawing and editing with visual foundation models},
  author={Wu, Chenfei and Yin, Shengming and Qi, Weizhen and Wang, Xiaodong and Tang, Zecheng and Duan, Nan},
  journal={arXiv preprint arXiv:2303.04671},
  year={2023}
}

@article{achiam2023gpt,
  title={Gpt-4 technical report},
  author={Achiam, Josh and Adler, Steven and Agarwal, Sandhini and Ahmad, Lama and Akkaya, Ilge and Aleman, Florencia Leoni and Almeida, Diogo and Altenschmidt, Janko and Altman, Sam and Anadkat, Shyamal and others},
  journal={arXiv preprint arXiv:2303.08774},
  year={2023}
}

@article{shen2023hugginggpt,
  title={Hugginggpt: Solving ai tasks with chatgpt and its friends in hugging face},
  author={Shen, Yongliang and Song, Kaitao and Tan, Xu and Li, Dongsheng and Lu, Weiming and Zhuang, Yueting},
  journal={Advances in Neural Information Processing Systems},
  volume={36},
  pages={38154--38180},
  year={2023}
}

@article{shinn2023reflexion,
  title={Reflexion: Language agents with verbal reinforcement learning},
  author={Shinn, Noah and Cassano, Federico and Gopinath, Ashwin and Narasimhan, Karthik and Yao, Shunyu},
  journal={Advances in Neural Information Processing Systems},
  volume={36},
  pages={8634--8652},
  year={2023}
}

@article{gou2023critic,
  title={Critic: Large language models can self-correct with tool-interactive critiquing},
  author={Gou, Zhibin and Shao, Zhihong and Gong, Yeyun and Shen, Yelong and Yang, Yujiu and Duan, Nan and Chen, Weizhu},
  journal={arXiv preprint arXiv:2305.11738},
  year={2023}
}

@article{madaan2023self,
  title={Self-refine: Iterative refinement with self-feedback},
  author={Madaan, Aman and Tandon, Niket and Gupta, Prakhar and Hallinan, Skyler and Gao, Luyu and Wiegreffe, Sarah and Alon, Uri and Dziri, Nouha and Prabhumoye, Shrimai and Yang, Yiming and others},
  journal={Advances in Neural Information Processing Systems},
  volume={36},
  pages={46534--46594},
  year={2023}
}

@article{liu2023agentbench,
  title={Agentbench: Evaluating llms as agents},
  author={Liu, Xiao and Yu, Hao and Zhang, Hanchen and Xu, Yifan and Lei, Xuanyu and Lai, Hanyu and Gu, Yu and Ding, Hangliang and Men, Kaiwen and Yang, Kejuan and others},
  journal={arXiv preprint arXiv:2308.03688},
  year={2023}
}

@article{xu2023rewoo,
  title={Rewoo: Decoupling reasoning from observations for efficient augmented language models},
  author={Xu, Binfeng and Peng, Zhiyuan and Lei, Bowen and Mukherjee, Subhabrata and Liu, Yuchen and Xu, Dongkuan},
  journal={arXiv preprint arXiv:2305.18323},
  year={2023}
}

@article{yao2023tree,
  title={Tree of thoughts: Deliberate problem solving with large language models},
  author={Yao, Shunyu and Yu, Dian and Zhao, Jeffrey and Shafran, Izhak and Griffiths, Tom and Cao, Yuan and Narasimhan, Karthik},
  journal={Advances in neural information processing systems},
  volume={36},
  pages={11809--11822},
  year={2023}
}

@article{wang2023plan,
  title={Plan-and-solve prompting: Improving zero-shot chain-of-thought reasoning by large language models},
  author={Wang, Lei and Xu, Wanyu and Lan, Yihuai and Hu, Zhiqiang and Lan, Yunshi and Lee, Roy Ka-Wei and Lim, Ee-Peng},
  journal={arXiv preprint arXiv:2305.04091},
  year={2023}
}

@article{kojima2022large,
  title={Large language models are zero-shot reasoners},
  author={Kojima, Takeshi and Gu, Shixiang Shane and Reid, Machel and Matsuo, Yutaka and Iwasawa, Yusuke},
  journal={Advances in neural information processing systems},
  volume={35},
  pages={22199--22213},
  year={2022}
}

@article{wei2022chain,
  title={Chain-of-thought prompting elicits reasoning in large language models},
  author={Wei, Jason and Wang, Xuezhi and Schuurmans, Dale and Bosma, Maarten and Xia, Fei and Chi, Ed and Le, Quoc V and Zhou, Denny and others},
  journal={Advances in neural information processing systems},
  volume={35},
  pages={24824--24837},
  year={2022}
}

@article{wang2022self,
  title={Self-consistency improves chain of thought reasoning in language models},
  author={Wang, Xuezhi and Wei, Jason and Schuurmans, Dale and Le, Quoc and Chi, Ed and Narang, Sharan and Chowdhery, Aakanksha and Zhou, Denny},
  journal={arXiv preprint arXiv:2203.11171},
  year={2022}
}

@article{li2023api,
  title={Api-bank: A comprehensive benchmark for tool-augmented llms},
  author={Li, Minghao and Zhao, Yingxiu and Yu, Bowen and Song, Feifan and Li, Hangyu and Yu, Haiyang and Li, Zhoujun and Huang, Fei and Li, Yongbin},
  journal={arXiv preprint arXiv:2304.08244},
  year={2023}
}

@article{qin2023toolllm,
  title={Toolllm: Facilitating large language models to master 16000+ real-world apis},
  author={Qin, Yujia and Liang, Shihao and Ye, Yining and Zhu, Kunlun and Yan, Lan and Lu, Yaxi and Lin, Yankai and Cong, Xin and Tang, Xiangru and Qian, Bill and others},
  journal={arXiv preprint arXiv:2307.16789},
  year={2023}
}

@article{li2023camel,
  title={Camel: Communicative agents for" mind" exploration of large language model society},
  author={Li, Guohao and Hammoud, Hasan and Itani, Hani and Khizbullin, Dmitrii and Ghanem, Bernard},
  journal={Advances in Neural Information Processing Systems},
  volume={36},
  pages={51991--52008},
  year={2023}
}

@inproceedings{park2023generative,
  title={Generative agents: Interactive simulacra of human behavior},
  author={Park, Joon Sung and O'Brien, Joseph and Cai, Carrie Jun and Morris, Meredith Ringel and Liang, Percy and Bernstein, Michael S},
  booktitle={Proceedings of the 36th annual acm symposium on user interface software and technology},
  pages={1--22},
  year={2023}
}

@article{wang2024survey,
  title={A survey on large language model based autonomous agents},
  author={Wang, Lei and Ma, Chen and Feng, Xueyang and Zhang, Zeyu and Yang, Hao and Zhang, Jingsen and Chen, Zhiyuan and Tang, Jiakai and Chen, Xu and Lin, Yankai and others},
  journal={Frontiers of Computer Science},
  volume={18},
  number={6},
  pages={186345},
  year={2024},
  publisher={Springer}
}

@book{kahneman2011thinking,
  title={Thinking, fast and slow},
  author={Kahneman, Daniel},
  year={2011},
  publisher={macmillan}
}

@inproceedings{wu2024autogen,
  title={Autogen: Enabling next-gen LLM applications via multi-agent conversations},
  author={Wu, Qingyun and Bansal, Gagan and Zhang, Jieyu and Wu, Yiran and Li, Beibin and Zhu, Erkang and Jiang, Li and Zhang, Xiaoyun and Zhang, Shaokun and Liu, Jiale and others},
  booktitle={First Conference on Language Modeling},
  year={2024}
}

@book{meyer1997object,
  title={Object-oriented software construction},
  author={Meyer, Bertrand},
  volume={2},
  year={1997},
  publisher={Prentice hall Englewood Cliffs}
}

@book{meyer2002design,
  title={Design by contract},
  author={Meyer, Bertrand},
  year={2002},
  publisher={Prentice Hall Upper Saddle River}
}

@inproceedings{hong2023metagpt,
  title={MetaGPT: Meta programming for a multi-agent collaborative framework},
  author={Hong, Sirui and Zhuge, Mingchen and Chen, Jonathan and Zheng, Xiawu and Cheng, Yuheng and Wang, Jinlin and Zhang, Ceyao and Wang, Zili and Yau, Steven Ka Shing and Lin, Zijuan and others},
  booktitle={The Twelfth International Conference on Learning Representations},
  year={2023}
}

@article{yang2025agrigpt,
  title={Agrigpt: A large language model ecosystem for agriculture},
  author={Yang, Bo and Zhang, Yu and Feng, Lanfei and Chen, Yunkui and Zhang, Jianyu and Xu, Xiao and Aierken, Nueraili and Li, Yurui and Chen, Yuxuan and Yang, Guijun and others},
  journal={arXiv preprint arXiv:2508.08632},
  year={2025}
}

@article{yang2025agrigpt-VL,
  title={Agrigpt-VL: Agricultural vision-language understanding suite},
  author={Yang, Bo and Chen, Yunkui and Feng, Lanfei and Zhang, Yu and Xu, Xiao and Zhang, Jianyu and Aierken, Nueraili and Huang, Runhe and Lin, Hongjian and Ying, Yibin and others},
  journal={arXiv preprint arXiv:2510.04002},
  year={2025}
}

@article{yang2025agrigpt-Omni,
  title={AgriGPT-Omni: A Unified Speech-Vision-Text Framework for Multilingual Agricultural Intelligence},
  author={Yang, Bo and Feng, Lanfei and Chen, Yunkui and Zhang, Yu and Zhang, Jianyu and Xu, Xiao and Aierken, Nueraili and Li, Shijian},
  journal={arXiv preprint arXiv:2512.10624},
  year={2025}
}

@article{xu2023tool,
  title={On the tool manipulation capability of open-source large language models},
  author={Xu, Qiantong and Hong, Fenglu and Li, Bo and Hu, Changran and Chen, Zhengyu and Zhang, Jian},
  journal={arXiv preprint arXiv:2305.16504},
  year={2023}
}

@article{guo2024stabletoolbench,
  title={Stabletoolbench: Towards stable large-scale benchmarking on tool learning of large language models},
  author={Guo, Zhicheng and Cheng, Sijie and Wang, Hao and Liang, Shihao and Qin, Yujia and Li, Peng and Liu, Zhiyuan and Sun, Maosong and Liu, Yang},
  journal={arXiv preprint arXiv:2403.07714},
  year={2024}
}

@article{shridhar2020alfworld,
  title={Alfworld: Aligning text and embodied environments for interactive learning},
  author={Shridhar, Mohit and Yuan, Xingdi and C{\^o}t{\'e}, Marc-Alexandre and Bisk, Yonatan and Trischler, Adam and Hausknecht, Matthew},
  journal={arXiv preprint arXiv:2010.03768},
  year={2020}
}

@inproceedings{ye2025toolhop,
  title={ToolHop: A Query-Driven Benchmark for Evaluating Large Language Models in Multi-Hop Tool Use},
  author={Ye, Junjie and Du, Zhengyin and Yao, Xuesong and Lin, Weijian and Xu, Yufei and Chen, Zehui and Wang, Zaiyuan and Zhu, Sining and Xi, Zhiheng and Yuan, Siyu and others},
  booktitle={Proceedings of the 63rd Annual Meeting of the Association for Computational Linguistics (Volume 1: Long Papers)},
  pages={2995--3021},
  year={2025}
}

@article{eiffert2022resource,
  title={Resource and response aware path planning for long-term autonomy of ground robots in agriculture},
  author={Eiffert, Stuart and Wallace, Nathan D and Kong, He and Pirmarzdashti, Navid and Sukkarieh, Salah},
  journal={Field Robotics},
  volume={2},
  pages={1--33},
  year={2022},
  publisher={FRPS}
}

@article{pal2022agricultural,
  title={An agricultural event prediction framework towards anticipatory scheduling of robot fleets: general concepts and case studies},
  author={Pal, Abhishesh and Das, Gautham and Hanheide, Marc and Candea Leite, Antonio and From, P{\aa}l Johan},
  journal={Agronomy},
  volume={12},
  number={6},
  pages={1299},
  year={2022},
  publisher={MDPI}
}

@article{agyeman2025semi,
  title={A semi-centralized multi-agent RL framework for efficient irrigation scheduling},
  author={Agyeman, Bernard T and Decardi-Nelson, Benjamin and Liu, Jinfeng and Shah, Sirish L},
  journal={Control Engineering Practice},
  volume={155},
  pages={106183},
  year={2025},
  publisher={Elsevier}
}

@article{escriba2024digital,
  title={Digital twins in agriculture: orchestration and applications},
  author={Escrib{\`a}-Gelonch, Marc and Liang, Shu and van Schalkwyk, Pieter and Fisk, Ian and Long, Nguyen Van Duc and Hessel, Volker},
  journal={Journal of agricultural and food chemistry},
  volume={72},
  number={19},
  pages={10737--10752},
  year={2024},
  publisher={ACS Publications}
}

@article{khosravi2025optimizing,
  title={Optimizing navigation and chemical application in precision agriculture with deep reinforcement learning and conditional action tree},
  author={Khosravi, Mahsa and Jiang, Zhanhong and Waite, Joshua R and Jones, Sarah E and Pacin, Hernan Torres and Singh, Arti and Ganapathysubramanian, Baskar and Singh, Asheesh Kumar and Sarkar, Soumik},
  journal={Smart Agricultural Technology},
  pages={101194},
  year={2025},
  publisher={Elsevier}
}

@article{chen2025multi,
  title={Multi-objective task allocation for electric harvesting robots: a hierarchical route reconstruction approach},
  author={Chen, Peng and Liang, Jing and Song, Hui and Qiao, Kang-Jia and Yue, Cai-Tong and Yu, Kun-Jie and Suganthan, Ponnuthurai Nagaratnam and Pedrycz, Witold},
  journal={arXiv preprint arXiv:2509.11025},
  year={2025}
}

@article{yang2025agripath,
  title={AgriPath: a robust multi-objective path planning framework for agricultural robots in dynamic field environments},
  author={Yang, Chenghan and Zheng, Dingkun and Chen, Siming and Mansurova, Madina and Belgibaev, Baurzhan and Zhao, Baidong},
  journal={Frontiers in Plant Science},
  volume={16},
  pages={1687747},
  year={2025}
}

@article{herron2025hierarchical,
  title={A Hierarchical Agentic Framework for Autonomous Drone-Based Visual Inspection},
  author={Herron, Ethan and Lee, Xian Yeow and Sin, Gregory and Diaz, Teresa Gonzalez and Farahat, Ahmed and Gupta, Chetan},
  journal={arXiv preprint arXiv:2510.00259},
  year={2025}
}

\end{document}